\documentclass[11pt]{article}

\usepackage[margin=1in]{geometry}
\usepackage[authoryear,round]{natbib}

\usepackage{amsmath,amsfonts,bm,amssymb}

\def\1{\bm{1}}

\DeclareMathAlphabet{\mathsfit}{\encodingdefault}{\sfdefault}{m}{sl}
\SetMathAlphabet{\mathsfit}{bold}{\encodingdefault}{\sfdefault}{bx}{n}

\newcommand{\R}{\mathbb{R}}

\newcounter{setting}[section]
\renewcommand{\thesetting}{\thesection.\arabic{setting}}

\newcounter{model}[section]
\renewcommand{\themodel}{\thesection.\arabic{model}}

\newcounter{initialization}[section]
\renewcommand{\theinitialization}{\thesection.\arabic{initialization}}

\usepackage[utf8]{inputenc} %
\usepackage[T1]{fontenc}    %
\usepackage{float}          %
\usepackage{hyperref}       %
\hypersetup{hidelinks}      %
\usepackage{url}            %
\usepackage{booktabs}       %
\usepackage{tabularx}       %
\usepackage{amsfonts}       %
\usepackage{nicefrac}       %
\usepackage{microtype}      %
\usepackage{xcolor}         %
\usepackage{placeins}       %
\usepackage{tcolorbox}

\usepackage{graphicx}

\title{Logical subspace in LLMs}

\author{Hope Kean \\ MIT \and Enric Boix-Adsera \\ UPenn}

\begin{document}

\maketitle

\begin{abstract}
Recent work has identified a human brain network specialized for abstract formal reasoning \citep{kean2025network}. Does the same hold true in language models? To answer this question, we introduce the minimal viable subspace (MVS) method, which searches
for the lowest-rank activation subspace at a layer that preserves task
performance when everything outside that subspace is ablated. Using MVS, we demonstrate low-rank subspaces supporting logical inference on Gemma and Qwen models. Furthermore, these subspaces exhibit a clear dissociation from model capacities on other tasks, such that retaining these late logic subspaces preserves inference while
impairing factual knowledge, working memory, cognitive control, and
arithmetic. Conversely, ablating them reduces logical inference accuracy to chance while
largely sparing these other capacities. Our results suggest a functionally localizable core machinery for logic akin to that in the human brain.

\end{abstract}

\section{Introduction}
\label{sec:introduction}

How does intelligence arise? Modern AI systems are typically evaluated by the range and difficulty of the problems they can solve. Yet behavioral competence alone does not reveal the internal organization that supports it. Diverse capacities could depend primarily on a broadly shared computational resource \citep{duncan2010}, on functionally specialized mechanisms \citep{kanwisher2010}, or on high-dimensional distributed representations from which task-relevant structure is dynamically selected \citep{fusi2016}. These possibilities are not wholly mutually exclusive, but they make different predictions about whether particular capacities can be causally isolated within a common architecture. Distinguishing them with respect to specific reasoning capacities is therefore central to understanding how intelligence is organized in both natural and artificial systems.

Logical inference provides one strong test case. The validity of an inference depends on its abstract relational structure rather than on the particular content assigned to its constituent variables: \emph{if $P$, then $Q$; $P$; therefore $Q$} remains valid regardless of what $P$ and $Q$ denote. The ability to extract such structure from its content is central to human reasoning \citep{rips2003, gentner1983structure, johnson2010mental} and has also been demonstrated, in controlled experiments, in large language models \citep{webb2023, lampinen2024,boixadsera2024}. Yet logic is only one component of what we think of as constituting intelligence, alongside language, memory, factual knowledge, quantitative reasoning, and other capacities. It therefore offers a tractable setting in which to ask whether a general-purpose transformer internally differentiates among cognitive functions.

A biological precedent comes from recent evidence that the human brain contains a network specialized for abstract formal reasoning, dissociable from both the language network and the domain-general Multiple Demand network \citep{kean2025network,kean2026language}. Recent work likewise suggests that LLMs recruit distinct neuron populations for broad cognitive domains, including language and executive, physical, and social reasoning \citep{han2026}. However, identifying decodable information or task-attributed neurons does not establish whether a capacity is implemented through a compact representation that is both sufficient for that capacity and selectively necessary for it. More generally, information can often be decoded from neural activity without being causally used by the system \citep{ritchie2019,kriegeskorte2019}.

\begin{figure}[t]
\centering
\includegraphics[width=0.92\textwidth, trim={2.6cm 15.5cm 2.8cm 0}, clip]{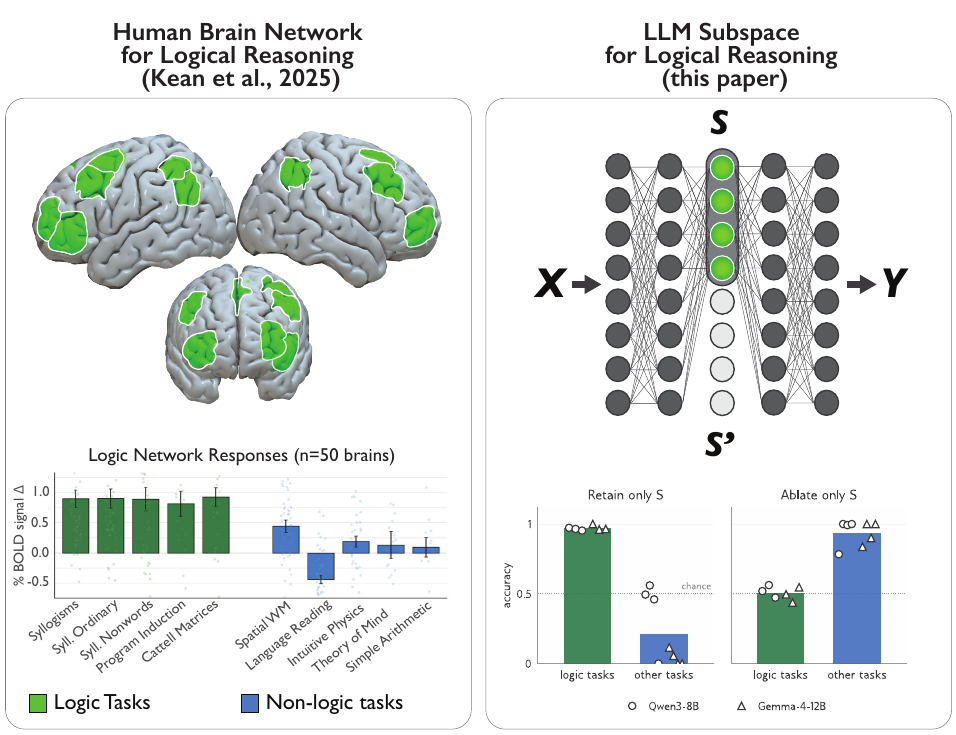}
\caption{\textbf{Left:} A human brain network for logical reasoning that selectively activates on logic tasks, but not on non-logic tasks \citep{kean2025network}. \textbf{Right:} This paper's approach to network functional localization, which finds ``logic subspaces'' of activations such that ablating these subspaces destroys logical capabilities but preserves control tasks, and retaining these subspaces preserves logical capabilities and destroys abilities on control tasks. We report the averages over Qwen3-8B (circles) and Gemma4-12B (triangles) for ablating subspaces at layers 22 and 21 respectively. Dashed orange lines indicate accuracy in the intact network.}\label{fig:teaser}
\end{figure}
In order to make progress on these questions, we introduce the \textbf{minimal viable subspace} (MVS) method for network localization. Our method searches for the lowest-rank activation subspace that preserves a model's performance on a target task when all orthogonal activity is removed. We identify MVSs across layers for one-step conditional inferences expressed using both ordinary and meaningless vocabulary. We then perform the complementary intervention---ablating only the identified subspace---to test whether it is selectively necessary for inference. Finally, we evaluate both interventions on language, working memory, factual knowledge, arithmetic, and a separate rule-based reasoning benchmark.

To foreshadow our results, we found that the smallest sufficient rank discovered at each layer falls sharply with depth, reaching
two dimensions in middle layers. Retaining this subspace preserves
near-ceiling logical performance while impairing control tasks; at middle layers,
ablating it reduces logic to near chance while largely sparing those
tasks. These effects generalize across ordinary and nonword arguments for the logical task. MVS thus provides
a causal approach to identifying compact, functionally differentiated
representations within a shared architecture, and understanding the representations underlying model capabilities.

\section{Related Work}
\label{sec:related-work}

\paragraph{Abstract reasoning in language models.}
Behavioral studies have shown that language models can perform forms of abstract relational and logical reasoning, while also exhibiting systematic effects of content and linguistic formulation \citep{webb2023,boixadsera2024,lampinen2024}. This work has firmly established models' behavioral competence in logical reasoning. More recently, \citet{yang2025emergent} have investigated the internal mechanisms supporting such abstract computations, identifying specialized attention heads supporting successive stages of symbolic abstraction, induction, and retrieval. In this work, we give a complementary view, demonstrating that the representation required to preserve logical function lies in a low-dimensional subspace.

\paragraph{Functional organization in LLMs.}
\citet{han2026} identify partially segregated neuron populations supporting broad cognitive domains and show that ablating neurons attributed to a domain selectively impairs tasks within that domain. \citet{gurnee2026} instead identify a sparse set of verbalizable representations (the J-space) with properties associated with a shared global workspace for reasoning. They find that this workspace supports explicit report and flexible inference, while text parsing and routine inference can proceed without it. Our analysis addresses a complementary level of organization: rather than localizing task-attributed units or a domain-general workspace, we ask whether a particular capacity can be isolated within a minimal low-dimensional activation subspace that is sufficient for that capacity but insufficient for others.

\paragraph{Causal subspaces and dictionary learning.}
Prior work has shown that low-rank subspaces of the residual stream can mediate communication between transformer components \citep{merullo2024talking}. \cite{yuan2026beyond} identify a subspace that is invariant to surface forms, indicating that some relevant variables for logical operations may lie in a low-dimensional subspace. Distributed alignment search (DAS) identifies subspaces corresponding to
prespecified causal variables through interchange interventions
\citep{geiger2024,wu2023interpretability,huang2024ravel}. In contrast, MVS is unsupervised. It searches for the smallest fixed subspace
that preserves performance on a specified task, without requiring
gold labels or prespecified causal variables. End-to-end sparse dictionary learning also preserves model
outputs \citep{braun2024}, but differs from MVS, which tests the sufficiency and
cross-task selectivity of a fixed low-rank subspace through
retention and ablation. At the level of attention heads, \citet{nam2025causal} learn gates that identify sparse sets of heads sufficient for a task. This asks a similar question to MVS, but over model components rather than residual-stream directions.

\section{Minimal viable subspaces: a method for localizing task machinery in LLMs}\label{sec:mvs}

We introduce the \textbf{\emph{minimal viable subspace} (MVS)} method for
localizing machinery that supports an LLM's performance
on a given task.
The method measures how well the model's task performance can
be preserved at a chosen subspace rank as follows. At layer $\ell$, let $h_\ell\in\mathbb{R}^d$ denote the residual-stream
activation, $\mu_\ell\in\mathbb{R}^d$ the mean activation over a
neutral corpus disjoint from all evaluated tasks, and
$U\in\mathbb{R}^{d\times k}$ a matrix whose orthonormal columns span
a learned rank-$k$ subspace. We project the activation onto this
subspace around the reference mean:
\begin{equation}
    h_\ell \;\mapsto\; \mu_\ell + UU^\top(h_\ell-\mu_\ell),
    \label{eq:retain}
\end{equation} This retains variation within the subspace
and removes variation orthogonal to it. We apply the projection at
every token position and let the remaining layers run normally.

Next, we learn the subspace $U$ by minimizing the KL divergence between the
unmodified and retained models' output distributions in order to keep the current behavior:
\begin{equation}
    \mathcal{L}_{\ell,\tau}(U)
    = \mathbb{E}_{x\sim\mathcal{D}_\tau}
    \left[
        \mathrm{KL}\!\left(
            p(\cdot\mid x)\,\big\|\,p^R_{\ell,U}(\cdot\mid x)
        \right)
    \right],
    \label{eq:objective}
\end{equation}
where $\mathcal{D}_\tau$ is the task's prompt distribution, and
$p$ and $p^R_{\ell,U}$ are the unmodified and retained models'
next-token distributions at the final prompt position.

We repeat this search across ranks and report how much held-out
task performance each optimized subspace retains. In this sense, MVS asks how
well a task can be supported within a subspace of a given rank. Applying the
method across layers shows how the relationship between rank and performance
changes with depth. More details are provided in Appendix~\ref{app:optim}.

\section{Experimental setup}
\label{sec:experimental-setup}

\paragraph{Models and layers.} We study Qwen3-8B ($36$ layers, $d=4096$) and
Gemma-4-12B ($48$ layers, $d=3840$), both in non-thinking mode. For each model,
we optimize separate subspaces at $15$ depths.

\paragraph{Optimization.} In order to optimize over the orthonormal matrix $U \in \R^{d\times k}$ that expresses the subspace, we instead optimize matrix $V\in\mathbb{R}^{d\times k}$ that is not necessarily constrained to be orthonormal. We obtain $U$ from $V$ at every step of the optimization by taking a QR
decomposition of $V$ and use its $k$ orthonormal columns as the basis $U$. At each model--layer--rank setting, with
the model weights frozen, we optimize $V$ using Adam for $1600$ steps, with
$16$ prompts per step and an initial learning rate of $10^{-2}$ cosine-annealed
to $\mathrm{lr}/20$. We test ranks $k\in\{1,2,4,8,16,32,64\}$ and run three
seeds for each model--layer--rank setting. For each seed, we take the median KL
over the final five logged steps, then average these medians across seeds. We
use this summary because single-minibatch KL estimates at adjacent late steps
can differ by an order of magnitude.

\subsection{Logical inference task}
\label{sec:logic-task}

Each logical-inference item contains a \emph{major premise}, a \emph{minor premise}, and a \emph{candidate conclusion} as in  \citet{aristotle_prior_analytics}. The model judges whether the conclusion
follows from the premises.

The full factorial design combines $2$ lexicons, $48$ contents, $14$ template
families, and $8$ argument forms, yielding
$2\times48\times14\times8=10{,}752$ items. We remove $32$ event-family items
in which a noun and verb share a stem. We remove both affected contents from both event families across
all $8$ forms ($2\times2\times8=32$). Thus, the final dataset contains $10{,}720$
items: $752$ in each event family and $768$ in each of the remaining $12$
families.

\paragraph{Argument forms.} For a rule $A\Rightarrow B$, the four standard
forms cross which proposition appears in the minor premise ($A$ or $B$)
with whether that proposition is affirmed or denied.

\begin{table}[H]
\centering\small
\caption{The four conditional-inference forms for a rule \(A \Rightarrow B\), crossing whether the minor premise concerns the antecedent or consequent with whether it affirms or denies that proposition. Modus ponens and modus tollens are valid; denying the antecedent and affirming the consequent are invalid.}
\label{tab:forms}
\begin{tabular}{llllc}
\toprule
Form & Fact cites & Fact & Conclusion & Validity \\
\midrule
Modus ponens (MP)             & antecedent $A$ & affirms & $B$        & valid   \\
Denying the antecedent (DA)   & antecedent $A$ & denies  & $\lnot B$  & invalid \\
Affirming the consequent (AC) & consequent $B$ & affirms & $A$        & invalid \\
Modus tollens (MT)            & consequent $B$ & denies  & $\lnot A$  & valid   \\
\bottomrule
\end{tabular}
\end{table}

These argument forms build on established paradigms for studying
conditional reasoning and deductive inference 
\citep[e.g.][]{johnson2002conditionals, monti2007functional}.

We instantiate each of the four argument forms with both an affirmative consequent ($A \Rightarrow B$) and a negated consequent ($A \Rightarrow \neg B$), yielding eight conditions. In both cases, MP and MT are valid, whereas AC and DA are invalid. Thus, consequent polarity is crossed with validity, yielding 2,680 items in each of the four polarity $\times$ validity combinations. Negation is expressed using wording appropriate to each template family: \textit{All trains are clean} becomes \textit{No trains are clean}, and \textit{a subset of} becomes \textit{disjoint from}, and \textit{always} becomes \textit{never}. Conditional templates use \textit{then it is not}.

\paragraph{Prompt.} Every item uses the same instruction. Only the three sentence syllogism
that follows changes. For full prompt examples across the different template
families, see Appendix~\ref{app:syl}.

\begin{tcolorbox}[
  colback=gray!4,
  colframe=gray!40,
  boxrule=0.4pt,
  arc=2pt,
  boxsep=2pt,
  left=4pt,
  right=4pt,
  top=3pt,
  bottom=3pt
]
\scriptsize
\begin{tabularx}{\linewidth}{
  @{}p{1.5cm}@{\hspace{0.8em}}>{\ttfamily\raggedright\arraybackslash}X@{}
}
\textsf{\textbf{Instructions}}
&
Please tell me whether the following reasoning is valid or invalid
(i.e., does the conclusion necessarily follow from the premises?).
Answer 1 if it is valid, or 2 if it is invalid.
Answer with just 1 or 2 and nothing else.
\\
\textsf{\textbf{Syllogism}}
&
The desk is pink.\newline
If the desk is cool, then it is pink.\newline
Therefore, the desk is cool.
\end{tabularx}
\end{tcolorbox}

\noindent
Labels are for illustration only;
the right column forms a single continuous prompt. The correct response is \texttt{2} (invalid):
the argument affirms the consequent.

\paragraph{Evaluation splits.}
Each evaluation split contains 256 items and tests
generalization along a different dimension:

\begin{itemize}
    \item \textsc{Items} uses unseen content within
    the 11 template families used during optimization.

    \item \textsc{Families} uses items from three
    template families excluded entirely from optimization, expressing set containment,
    hypotheticals, and quantification. Each family instantiates all four argument
    forms with both affirmative and negated consequents.

    \item \textsc{Nonwords} uses the training template
    families with a pronounceable nonword lexicon,
    removing any possible real-world plausibility confounds.
\end{itemize}

We call this logical inference task the \textit{logic localizer}.

\subsection{Control tasks and scoring}
\label{sec:control-tasks}

To assess \textit{selectivity}, we evaluate the effects of the
same interventions on tasks spanning memory, language, reasoning,
world knowledge, cognitive control, and arithmetic. These include
word recall; a sentence-versus-pseudoword-list task adapted
from the human language localizer
\citep{fedorenko2011functional}; multi-step deduction over invented
categories; MMLU questions \citep{hendrycks2020measuring};
a Stroop variant \citep{stroop1935studies} and an adaptation
of the Multi-Source Interference Task
\citep{bush2003multi}; Raven-style matrix completion;
and multiplication. We additionally assess preservation of
next-token prediction with a \textit{language modeling task}, described below. For full task definitions, example
prompts, and accuracies without intervention
see Appendix~B.2.

\paragraph{Response formats and scoring.}
Forced-choice tasks use the same response labels as the logic
localizer (\texttt{1} and \texttt{2}), except for multi-source
interference and the alternative MMLU formats.
The meaning of each label is specified in the task prompt.
Multi-source interference uses \texttt{A} and \texttt{B}
as response labels to distinguish the labels from the digit
answers. Multiplication requires the model to produce the
numerical answer directly. For forced-choice tasks, we score
the first generated character against the correct response
label; responses outside the permitted label set are counted
as incorrect.

\paragraph{Response-label controls.}
The shared \texttt{1}/\texttt{2} format raises the possibility
that the isolated subspace merely preserves the response-label
mapping rather than the task computations of interest.
If intervention effects were due only to a particular
response-label mapping, they should change when the labels
are changed while the questions and answer choices are
held fixed. We therefore evaluate the same binary MMLU
items using either \texttt{1}/\texttt{2} or
\texttt{A}/\texttt{B} labels. Without intervention, accuracy
differs between the two formats by at most 0.6 percentage
points for either model, providing closely matched baselines
for this comparison. To test whether the effects extend
beyond binary choice, we additionally evaluate Qwen on MMLU
in its standard four-option format. Finally, multiplication
tests whether the effects extend to producing numerical
answers without a fixed set of response labels.

\paragraph{Language modeling task}

To assess whether an intervention preserves fundamental \textit{language modeling}
behavior, we also compare the next-token distributions of the
unmodified and intervened models over a WikiText-2 continuation
of more than 90 words. We score
the task using the mean per-token KL divergence. Lower values
indicate better preservation of the unmodified model's
predictions; zero indicates identical distributions.
This measure captures changes across the full output
distribution, including those that leave the most likely
token unchanged.

\section{Results}

Across Qwen3-8B and Gemma4-12B, we \textbf{(1)} show that  the MVS method can be used to find low-rank
activation subspaces such that logical inference capabilities are preserved
when all components outside the subspace are ablated. 
Next, we find that these subspaces \textbf{(2)} reveal a dissociation between logic and other capabilities, since ablating these subspaces at intermediate layers selectively disrupts logical
inference while preserving performance on other tasks such as general language modeling.

Together, these results provide evidence for
a functionally specialized machinery that is both necessary and sufficient for logical inference
in LLMs.

\subsection{Low-rank activation subspaces that are sufficient for logical inference}

\begin{figure}[t]
\centering
\includegraphics[width=\textwidth, trim={0cm 17.2cm 0cm 0.5cm,clip}]{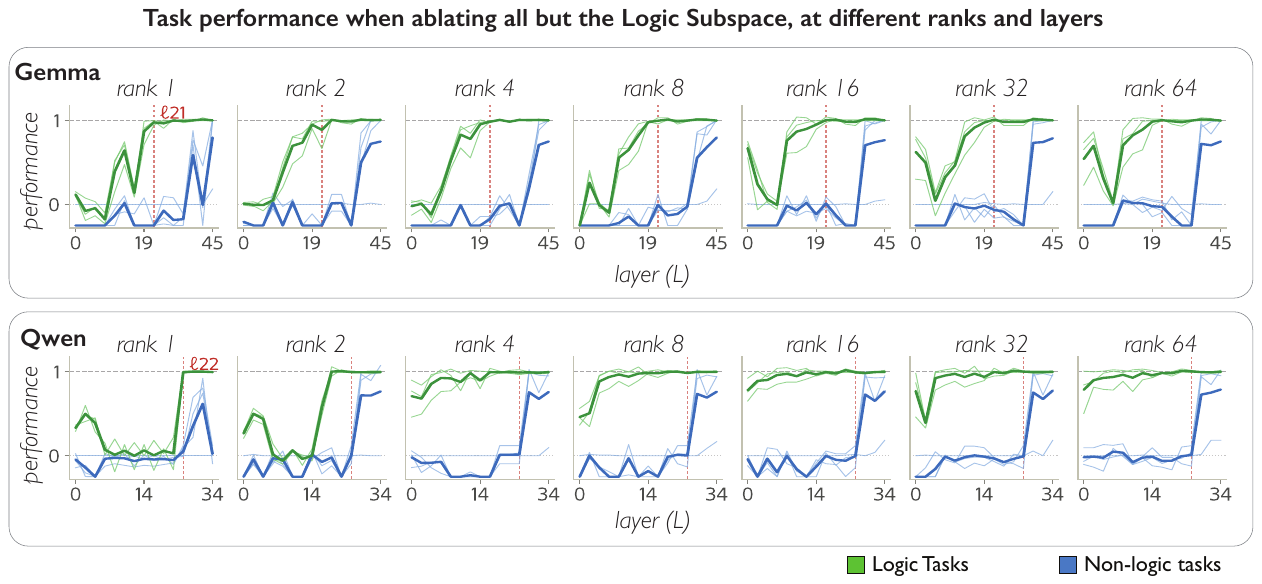}
\caption{
Task performance on Gemma4-12B and Qwen3-8B when retaining only the subspace found with MVS on logic using the projection \eqref{eq:retain}. Our measure of performance is $(\text{retained accuracy}-\text{chance})/(\text{intact accuracy}-\text{chance})$. A score of 1 indicates the fully intact model's capabilities are preserved, while a score of 0 indicates chance-level accuracy. Note that logic tasks (including held-out formulations) are preserved by projection to the low-rank subspaces, but non-logic tasks are not (until the final layers, which correspond to a shared output channel between logic and the other tasks). This figure with fine-grained task info is available in Appendix~\ref{app:extra-results}.}\label{fig:task-by-layer-rank}
\end{figure}

\paragraph{Low-rank subspaces suffice for logical inference.} We apply the MVS method at various layers to find subspaces that preserve the model's behavior on the syllogisms task under the projection operation \eqref{eq:retain}. Figure~\ref{fig:task-by-layer-rank} reports Qwen3-8B and Gemma-4-12B's accuracy across tasks, when retaining only the optimized logic subspace at different layers.

\paragraph{The smallest sufficient rank decreases with depth.}
The rank of the subspace needed to preserve performance on logical reasoning tasks  generally decreases with depth. 
The rank required to preserve logical inference decreases
with depth in both models (Figure~\ref{fig:task-by-layer-rank}). At later layers,
retaining only a small fraction of the activation space
preserves performance on the logic task, even as performance
on other tasks deteriorates. Thus, these subspaces preserve
logical inference without preserving the models' broader
capabilities.

\begin{figure}[h]
\centering
\includegraphics[width=\textwidth,trim={0cm 9cm 0cm 9cm, clip}]{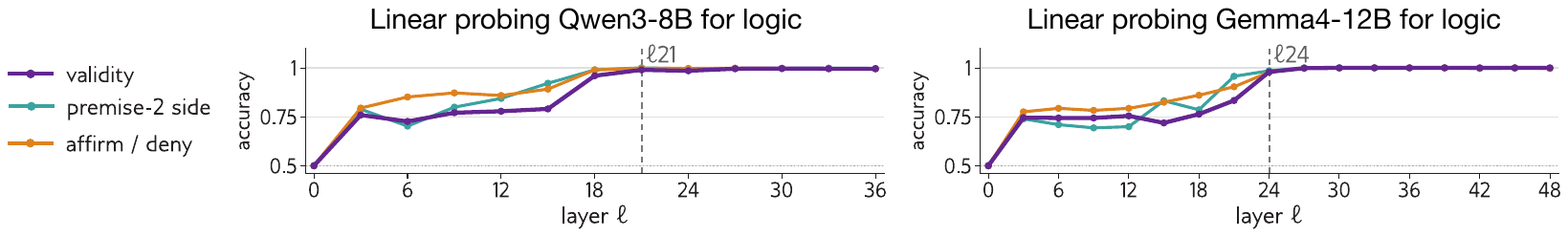}
\caption{Probing each layer of Qwen3-8B and Gemma4-12B on the syllogistic reasoning tasks shows that logical features become linearly decodable with depth, with near-perfect decodability by layer 21 and layer 24, respectively.}\label{fig:linear-probes}
\end{figure}

This converges with the fact that information about argument structure and validity becomes
increasingly accessible to linear probes across the middle
layers of both models {\color{red}(Figure~\ref{fig:linear-probes})}. Probes decode which
proposition the second premise concerns, whether it affirms
or denies that proposition, and whether the argument is
valid. Decoding reaches near-perfect accuracy by layer~21
in Qwen3-8B and layer~24 in Gemma4-12B. 

\paragraph{The effects are invariant to lexical content.}
Although MVS trains to preserve logical syllogism performance where the syllogisms contain English word arguments, the effects of ablating or retaining it generalize to logical syllogisms on nonword arguments, as well as held-out formulations of logical syllogisms; see task definitions in Section~\ref{sec:logic-task}. The effect is therefore not tied to particular lexical items and is consistent with sensitivity to abstract inferential structure.

\subsection{Retaining and ablating logic subspaces selectively dissociates logical inference from other tasks, such as language modeling}

\paragraph{A late MVS is selectively necessary and sufficient.}
In addition to retaining only the logic subspace, we also consider ablating it, replacing $UU^\top$ in
\eqref{eq:retain} with $I-UU^\top$. Retention tests whether the subspace is sufficient for
logical-inference performance. Ablation of the same subspace
then tests whether its removal selectively impairs logical
inference while sparing the tested control tasks. As we report in the right panel of Figure~\ref{fig:teaser}, at layer 22 of Qwen3-8B a rank-2 MVS and at layer 21 of Gemma4-12b a rank-1 MVS appear to be logic-specific. These are subspaces around where the linear probes indicate that validity of the logical syllogism first becomes linearly probeable. Ablating only this subspace reduced conditional-inference accuracy to approximately chance, while largely preserving performance on MMLU, working memory, a Stroop task variant, and multiplication. Conversely, retaining only the MVS preserved logic while substantially impairing these control tasks.
\begin{figure}
\centering
\includegraphics[width=\textwidth]{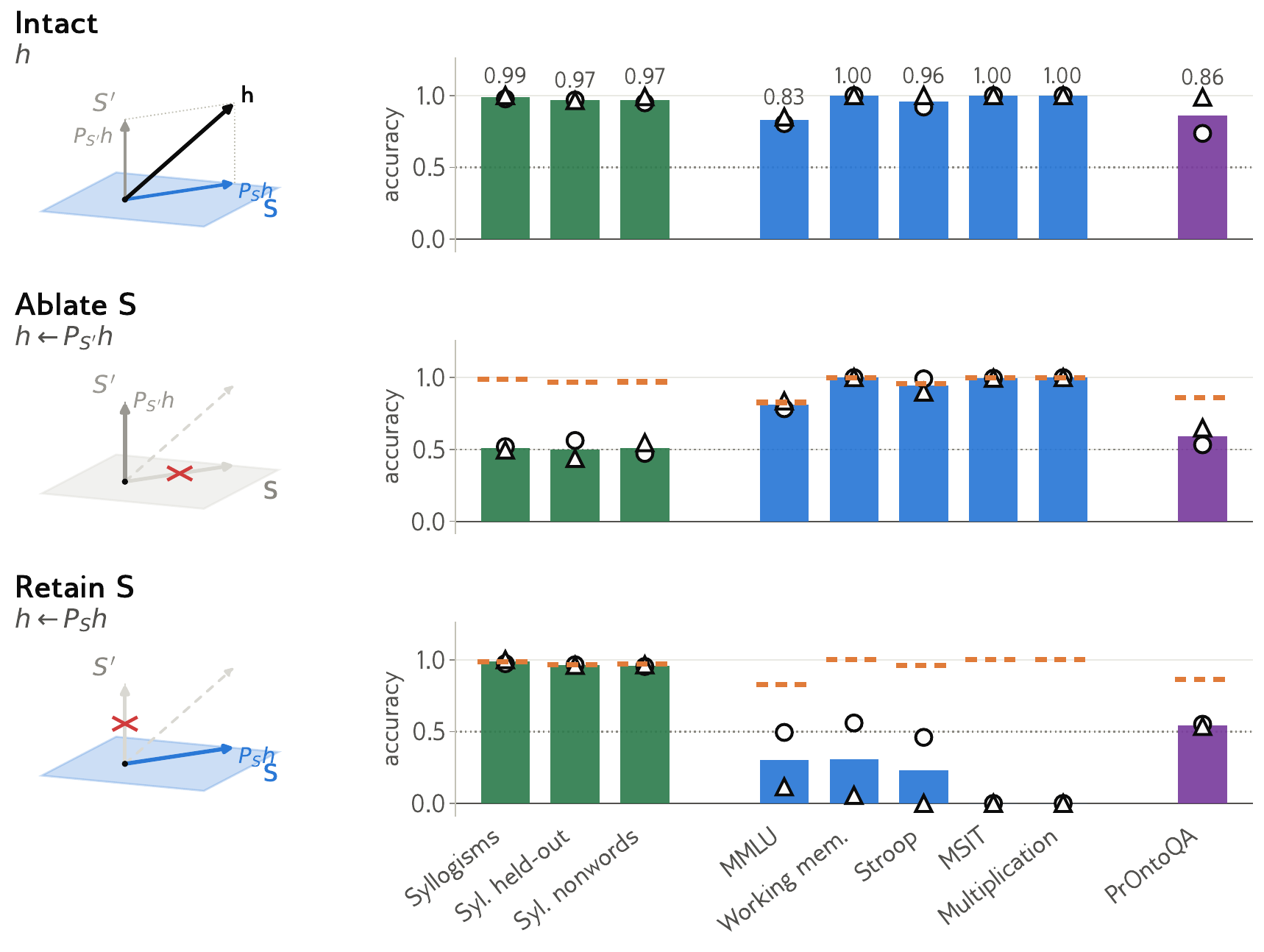}
\caption{\textbf{A low-rank subspace is selectively necessary and sufficient
for logical inference.} Qwen3-8B was modified at $\ell_{22}$ with rank $2$
and Gemma-4-12B at $\ell_{21}$ with rank $1$; both use the $U$ learned for
syllogisms by Eq.~(2). \textbf{Top:} the intact model. \textbf{Middle:}
ablating $S$, $h_\ell \mapsto \mu_\ell + (I - UU^{\top})(h_\ell - \mu_\ell)$.
\textbf{Bottom:} retaining $S$, Eq.~(1). Bars are the mean over the two
models and individual points indicate individual models ($\circ$ Qwen3-8B, $\triangle$
Gemma-4-12B); orange dashes give each task's intact accuracy. Ablating $S$ drops all three syllogism sets to chance
($0.99 \rightarrow 0.51$, $0.97 \rightarrow 0.50$, $0.97 \rightarrow 0.51$)
while factual knowledge, working memory, cognitive control and arithmetic
performance is unchanged. Retaining $S$ reverses the effect, such that syllogisms are
preserved ($0.99$, $0.96$, $0.96$) and the controls fall to or below chance. PrOntoQA, considered to be a multi-step logic/deduction task, performance is affected by \textit{both} interventions and is plotted separately, we return to this in the Discussion.}
\label{fig:dissociation}
\end{figure}

\paragraph{When other tasks emerge }Additionally, until the final layers, retaining only the logic subspace leads to significantly impaired performance on other tasks. For these other tasks, the different capacities become supportable by compact subspaces at different depths: for Qwen3-8B under rank-\(2\) logic subspace retentions the language versus nonwords classification task is first recovered at layer 26, and working memory at layer 30. 

\begin{figure}
    \centering
    \includegraphics[width=0.65\textwidth]{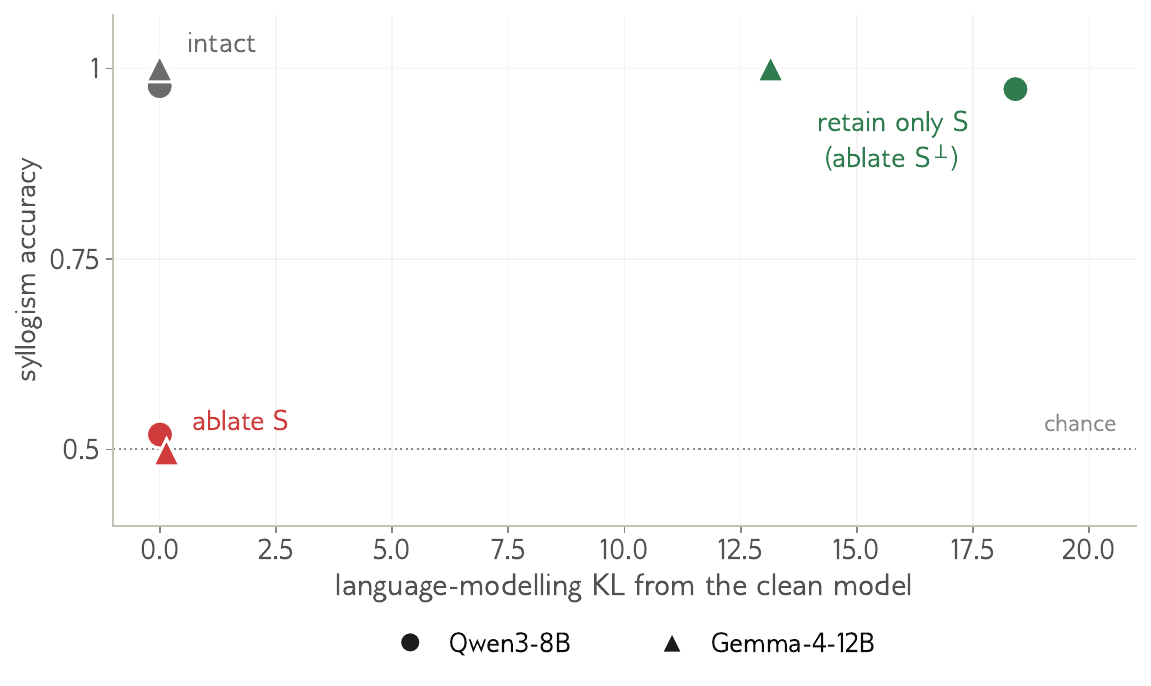}
    \caption{Ablating the logic subspaces at intermediate layers of Gemma and Qwen preserves language modeling ability but destroys logic. Retaining the subspaces preserves logic, but destroys language modeling ability. This shows the dissociation results presented in Figure~\ref{fig:teaser} extend to language modeling, which does not have an accuracy score from 0 to 1 so it was not presented in that figure.}
    \label{fig:syl_vs_lm}
\end{figure}

\paragraph{Logical inference is dissociable from general-purpose language modeling.}
We evaluated language modeling using mean per-token KL divergence
between the unmodified and intervened models' next-token distributions
over text continuations, rather than accuracy; see task definition in Section~\ref{sec:control-tasks}. At the selected
points (Qwen: layer 22, rank 2; Gemma: layer 21, rank 1), retaining the
logic subspace preserves near-baseline syllogism accuracy while
substantially increasing continuation KL. Conversely, ablating the
same subspace reduces syllogism accuracy to chance while leaving
continuation KL close to zero. Thus, logical judgments can survive
substantial disruption of language-model predictions, and can be
impaired while those predictions remain largely preserved. See Figure~\ref{fig:syl_vs_lm}

\subsection{Visualizing the logic subspace}

Finally, we visualize the rank-2 logic subspace at the intermediate layers identified in Figures~\ref{fig:dissociation} and \ref{fig:syl_vs_lm}, showing that in this subspace the networks contain cluster structure that matches the abstract argument form of the syllogism.

\begin{figure}[h]
    \centering
    \includegraphics[width=0.75\textwidth]{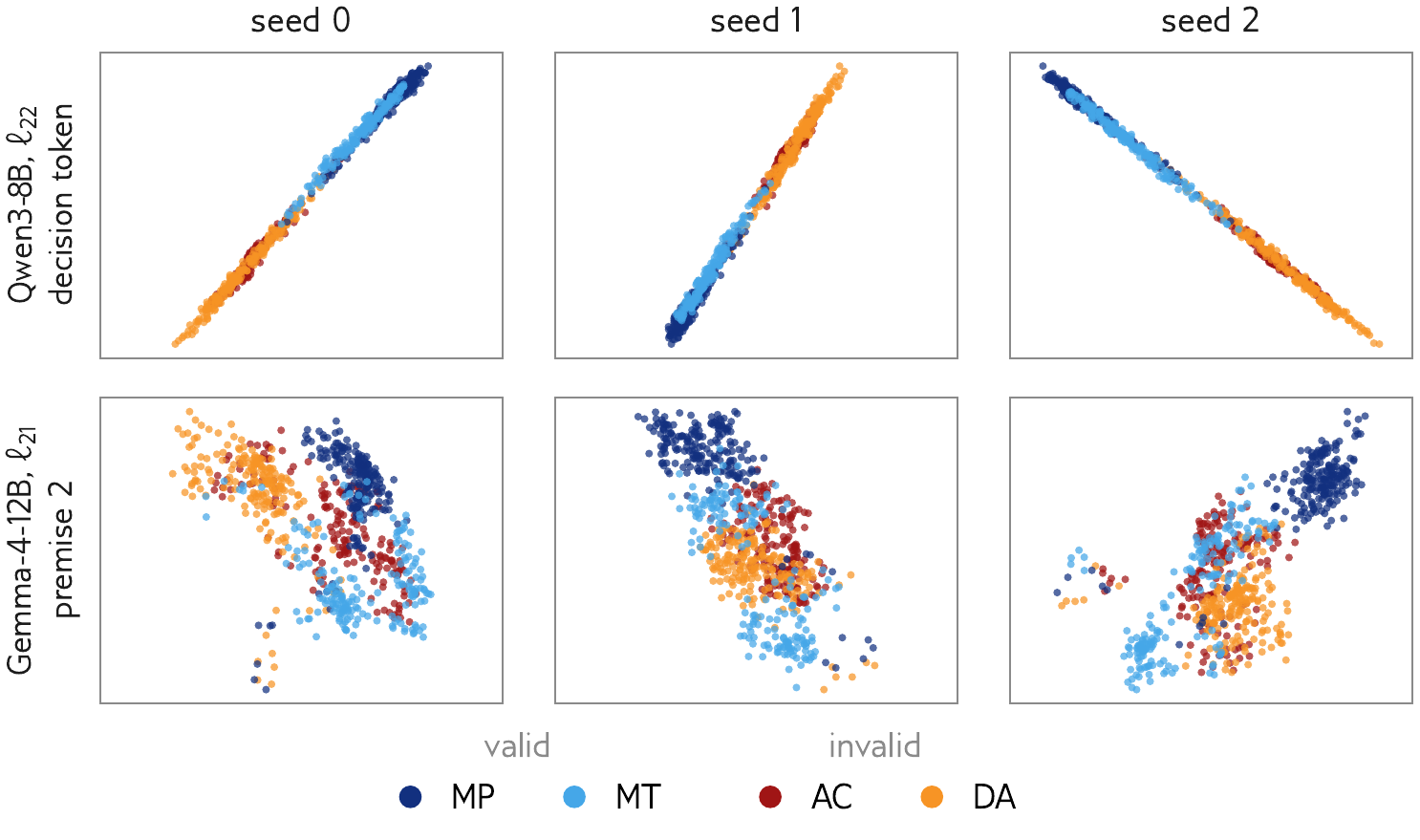}
    \caption{We plot the projection to the rank-2 logic subspace of the activations at Layer 22 of Qwen3-8B and Layer 22 of Gemma4-12B of syllogistic statements of various forms: Modus Ponens (MP), Denying the Antecedent (DA), Affirming the Consequent (AC), and Modus Tollens (MT). The activations for Qwen3-8B are linearly separable by validity. The activations for Gemma4-12B show cluster structure corresponding to the different logical forms. More plots are available in Figure~\ref{fig:logic-subspace-visualization}.}
\end{figure}

\section{Discussion}
\label{sec:discussion}

Our results demonstrate that logical inference capabilities can be retained while ablating all but a very low-rank subspace of a layer's activations. Additionally, ablating just the logic subspace on certain layers preserves the model's capabilities on reference control tasks. These results demonstrate a strong dissociation between the machinery that LLMs use for different tasks. This adds to substantial evidence for localized representations
in language models, including sparse feature decompositions
\citep{braun2024}, causal subspaces identified by distributed alignment
search \citep{geiger2024}, and domain-selective neuron populations
\citep{han2026}. 

Additionally, the subspaces we recover appear to extend beyond particular lexical contents, since they work with several templates of syllogisms and generalize to these templates being filled with nonwords as well as with words.

The retention procedure connects to pruning and the lottery ticket
hypothesis: both ask how much of the network can be removed while preserving the
potential for successful behavior 
\citep{frankle2019lottery,hinton2015distilling,caruana}. We demonstrate that this pruning can be quite significant, removing all but few dimensions at the subspace level, while retaining performance on logical inference tasks.

MVS opens the way to several interesting future directions of investigation, which we discuss below.

\paragraph{A cartography of tasks in LLMs via MVS}
Applying the minimal viable subspaces (MVS) to tasks beyond logical inference could reveal which capabilities depend on
shared directions and which admit distinct sufficient subspaces. 

\paragraph{Comparison with PCA and the J-space}
Although MVS starts within the leading principal subspace of the activation covariance, it is trained to preserve task behavior rather than explained variance and is free to move away from its initialization. It would be useful to test whether a task-selected subset of principal components can match the sufficiency and selectivity of the learned MVS. This would show whether covariance structure already supplies the relevant directions or whether task-specific optimization is needed.

A second comparison is with the J-space of \citet{gurnee2026}. The J-space supports flexible inference but is not needed for several forms of routine processing. Comparing the logic MVS with the J-lens directions active during the task would test whether the logical inference studied here uses that shared workspace or a separate route.

\paragraph{Interpretable variables in the MVS} Comparing MVS with DAS subspaces aligned to known logical variables
could further connect the preservation of a behavior to the particular
inferential distinctions that support it. Since MVS is unsupervised, additionally it would be of interest to automatically annotate or discover any interpretable latent variables living in the recovered subspaces.

\paragraph{Matched controls and task-general structure} The research program espoused by this work follows the
broader logic of functional localization in neuroscience, in which one identifies a
candidate system and characterizes its function profile. In neuroscience, this is often done through carefully
chosen contrasts \citep{kanwisher2010}, in which the conditions are matched except in one dimension of interest. Taking inspiration from neuroscience, matched tasks could potentially allow us to isolate general machinery from task-specific machinery across a very wide suite of operations. Additionally, subtracting out a ``core'' machinery of MVSs from different tasks from the MVS for logic, might yield subspaces at \textit{all} layers that, when ablated, destroy only logic performance.

\paragraph{Models in thinking mode} We deliberately studied non-thinking modes to isolate the model when evaluating a single conditional inference, which we view as an atomic
component of reasoning.
Extending MVS to models generating extended chains of thought would
require a larger compute budget than the present study and would allow
us to ask how sufficient subspaces change across reasoning steps.
Geometric studies already relate logical structure and stages of
reasoning to representation trajectories
\citep{zhou2026geometry,sun2026trajectories}. Combining these analyses
with MVS could test whether successive steps reuse a shared subspace,
recruit distinct subspaces, or depend on their interaction as elementary
inferences are composed into longer arguments.

\FloatBarrier

\subsection*{Acknowledgements}
The authors would like to thank Taylor Webb, Andrea de Varda, Ev Fedorenko, and Faith Kean for engaging discussions and helpful comments.

\subsection*{AI use statement}

Separately from the language models evaluated in this study, we used generative AI tools to assist with implementing code for the experiments and generating the synthetic syllogism dataset. We also used these tools to retrieve and discover relevant literature and to draft and polish portions of the manuscript where appropriate. Any AI outputs were heavily edited by the authors. We did not use generative AI to formulate or refine the research hypotheses, develop the methodology, design or provide feedback on the experiments, or interpret the results. The authors take full responsibility for the final content of this work, including its text, claims, code, and data.

\bibliography{bibliography.bib}
\bibliographystyle{plainnat}

\clearpage
\appendix

\section{Further results}\label{app:extra-results}

Here we provide further analysis with respect to the results reported in the main text, as well as a version of Figure~\ref{fig:task-by-layer-rank} containing the fine-grained detail of performance of the different tasks under retentions and ablations of the recovered logical subspaces of different ranks.

\begin{figure}[H]
    \centering
    \includegraphics[width=\textwidth]{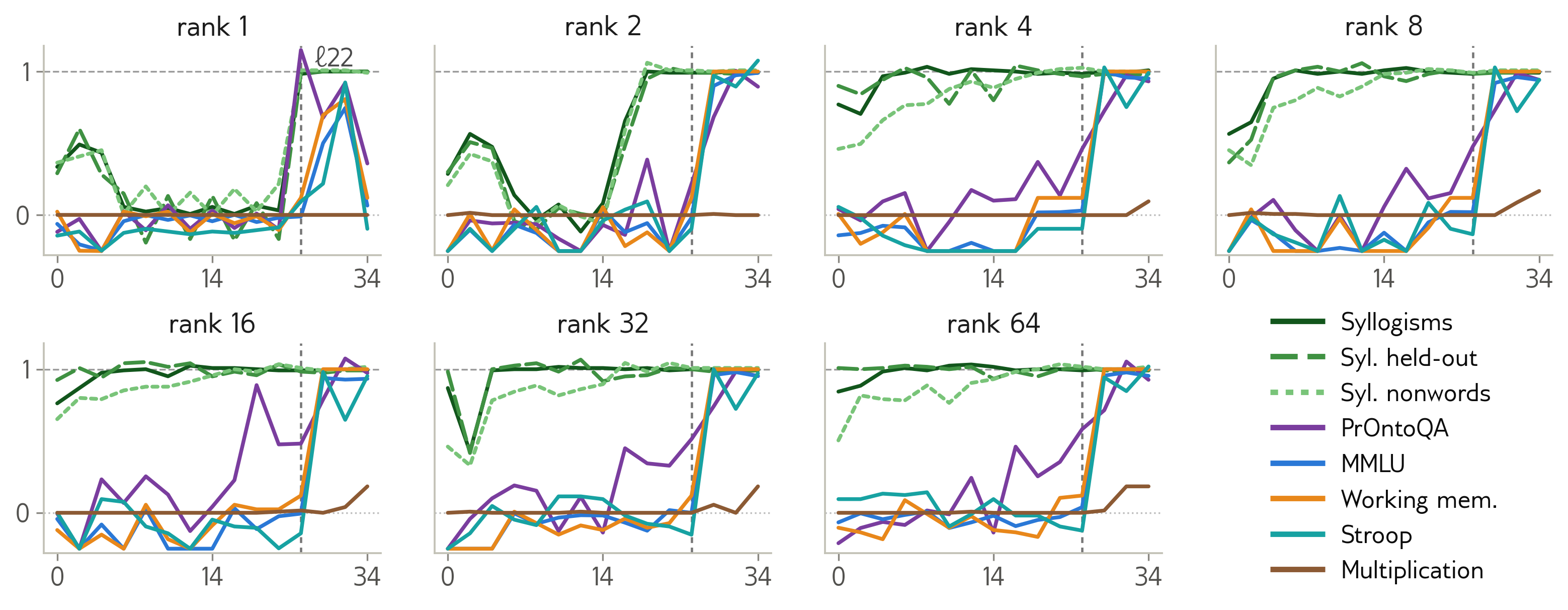}
    \caption{Full results from Figure~\ref{fig:task-by-layer-rank} for performance recovered on Qwen3-8B when retaining logic subspace at different ranks and layers.}
\end{figure}

\begin{figure}[H]
    \centering
    \includegraphics[width=\textwidth]{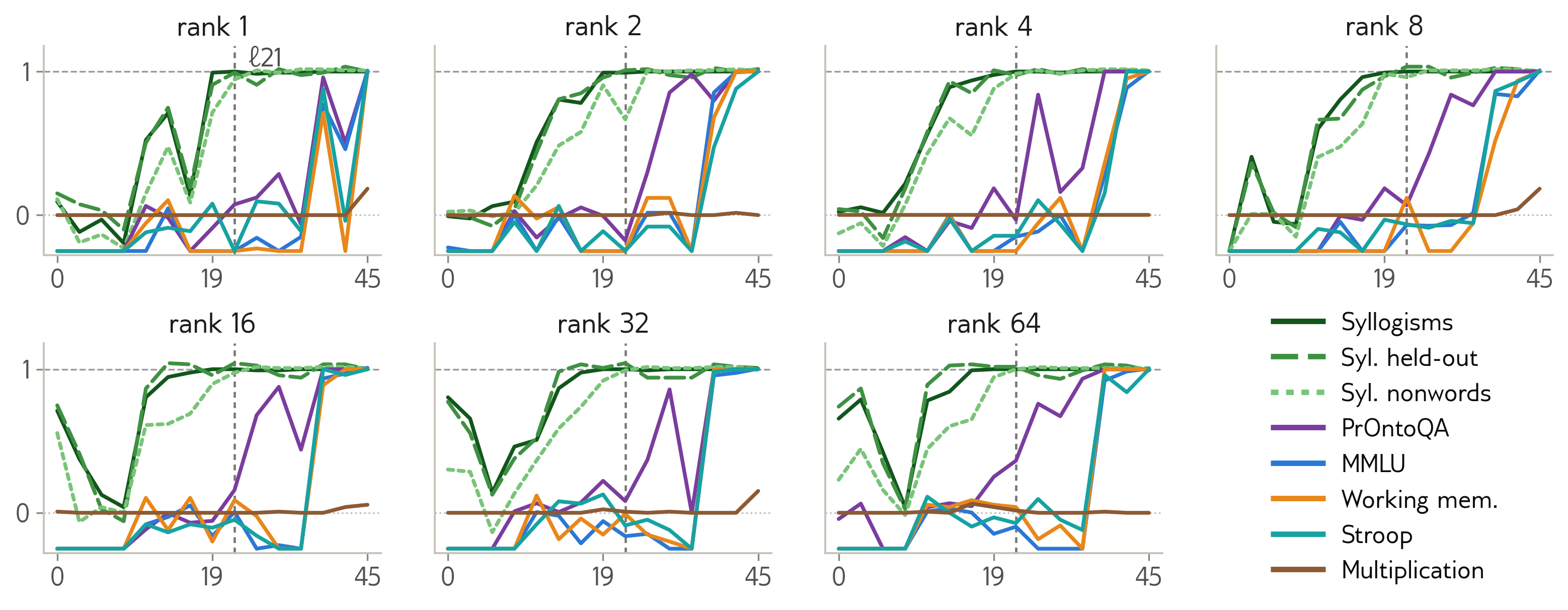}
    \caption{Full results from Figure~\ref{fig:task-by-layer-rank} for performance recovered on Gemma4-12B when retaining logic subspace at different ranks and layers.}
\end{figure}

\begin{figure}[t]
\centering
\begin{minipage}[t]{0.49\textwidth}
\centering
\includegraphics[width=\linewidth]{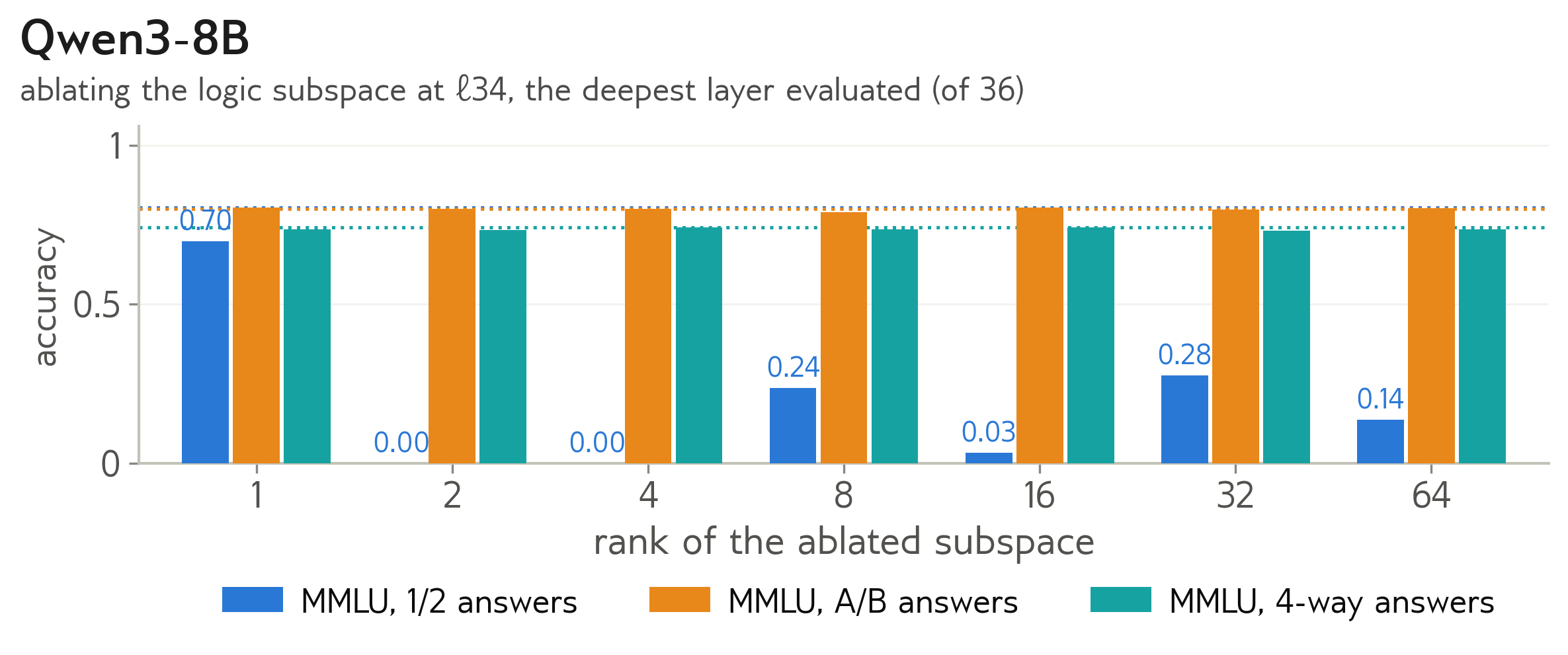}
\end{minipage}\hfill
\begin{minipage}[t]{0.49\textwidth}
\centering
\includegraphics[width=\linewidth]{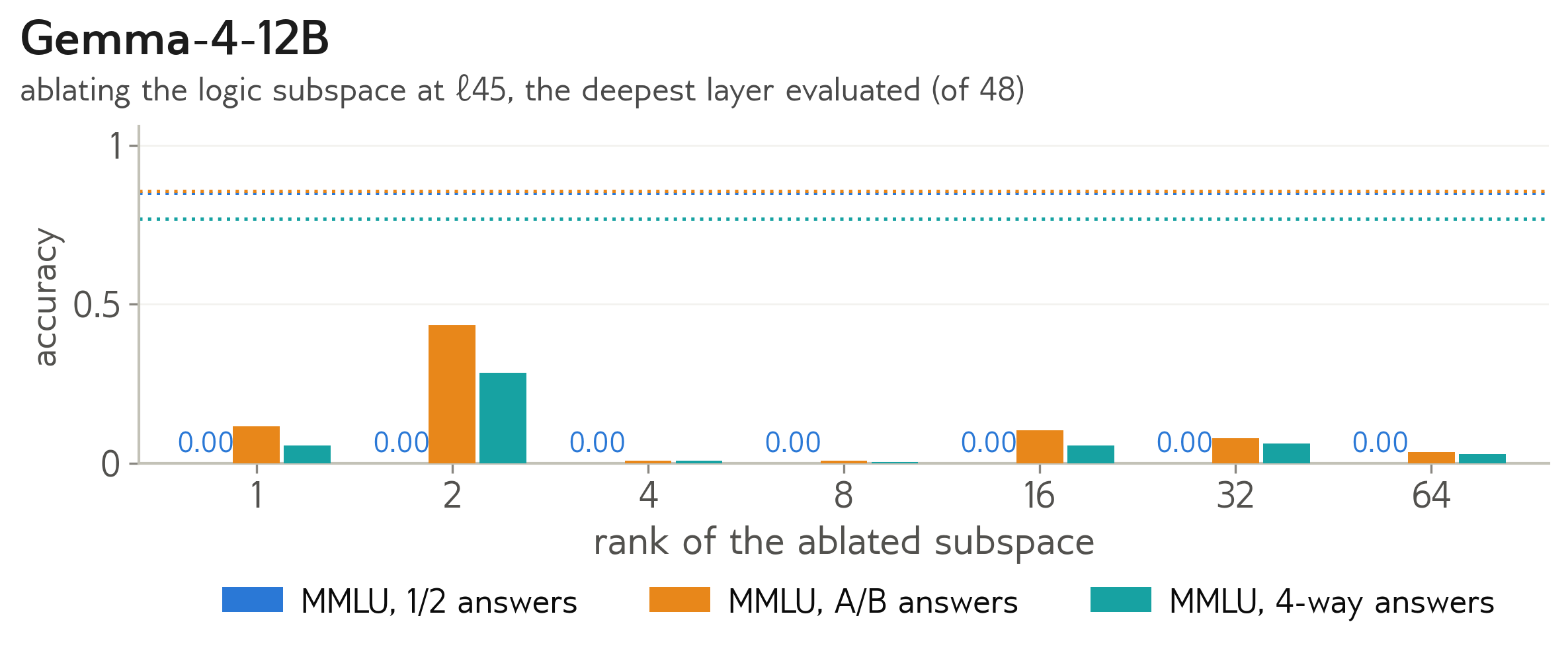}
\end{minipage}
\caption{Left (Qwen3-8B): The rank-2 logic subspace ablation of the last evaluated layer on Qwen hurts performance on MMLU with 1/2 output format, but not on MMLU with A/B output format. The percentage of correctly outputted answers for MMLU 1/2 drops to 0\% for that layer, indicating that this subspace controls the output channel for the logic task, which is shared with the output channel for MMLU 1/2 but not for MMLU A/B. Right (Gemma-4-12B): At the last evaluated layer of gemma, the effect of ablating the logic subspace still shows a contrast between MMLU 1/2 and MMLU A/B, although it is not as dramatic as for Qwen in the left panel. The general decrease in accuracy on all formats might hypothetically correspond to a more abstract output channel for Gemma than for Qwen, which is shared across these formats.}\label{fig:mmlu_ab_vs_mmlu_12}
\end{figure}

\paragraph{Late-layer subspace on Qwen acts as an output channel for 1 versus 2} Our formulation of the syllogism task asks for the model to output 1 if it is valid, and 2 if it is invalid. Similarly, when feasible, we formulate our control tasks to require a 1/2 output channel (see task definitions in Section~\ref{sec:control-tasks}). Here we observe that in the last layer of the network for which we compute the logic subspace (well after the syllogism validity is linearly probeable), the subspaces that preserve performance on the syllogisms task seem to be dedicated to the output channel machinery for this task. As evidence, we note in Figure~\ref{fig:mmlu_ab_vs_mmlu_12}, that in the final layer of Qwen ablation of the subspaces for logic hurts a binary-choice version of MMLU with a 1/2 output channel, but leaves the binary-choice version of MMLU with a A/B output channel unaffected.

\begin{figure}
\centering
    \includegraphics[width=0.8\textwidth]{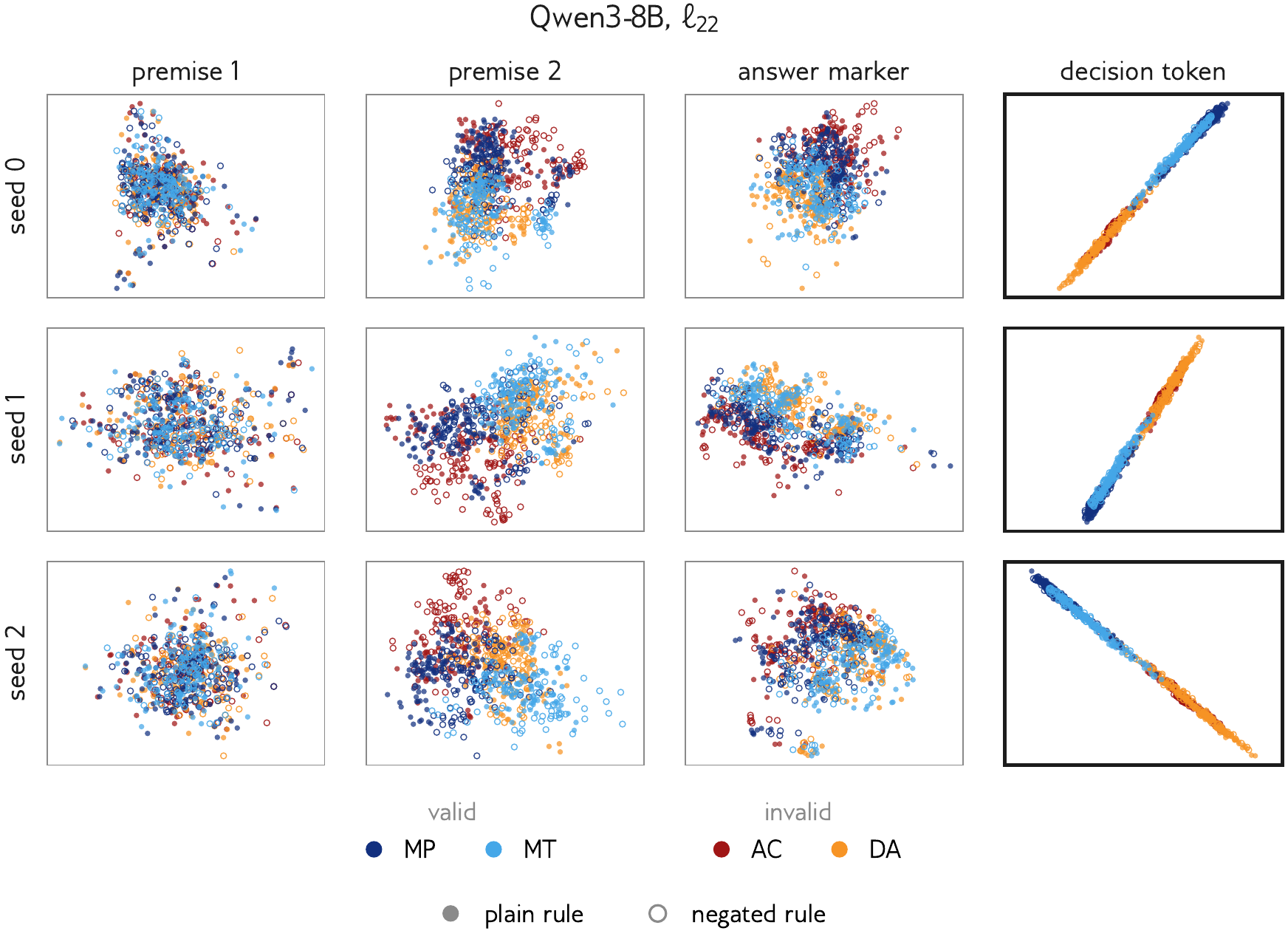}
    \includegraphics[width=0.8\textwidth]{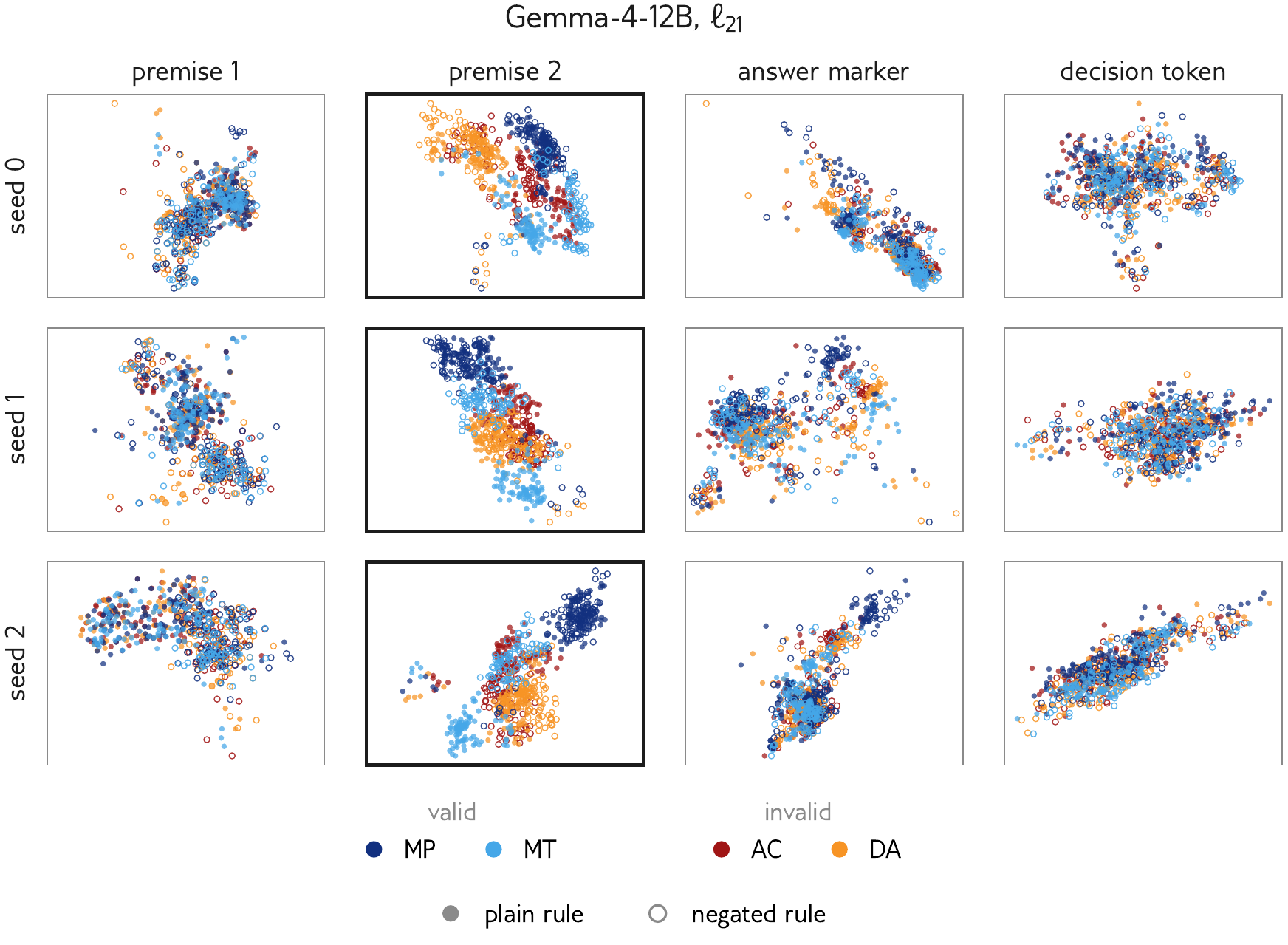}
    \caption{Visualization of projection of the activations of Qwen3-8B and Gemma4-12B to the rank-2 logic subpaces at layer 22 and layer 21, respectively. We see that cluster structure emerges, corresponding to the different abstract underlying syllogisms.}\label{fig:logic-subspace-visualization}
\end{figure}

\paragraph{Visualizing the logic subspaces} Figure~\ref{fig:logic-subspace-visualization} extends the main-text visualization by showing the projections at four token positions and across three optimization seeds for each model.

\FloatBarrier
\section{Tasks}\label{app:tasks}

\subsection{Logical inference}
\label{app:syl}

\paragraph{Template families.} Table~\ref{tab:families} lists all $14$,
grouped by the logical formulation of the rule. Premise order is a property of
the family rather than an independent factor: \textsc{fact\_first} states the
particular fact before the rule, and all others state the rule first. Two
further surface factors vary within family: the conclusion marker
(\emph{Therefore,} / \emph{Hence,} / \emph{Thus,} / \emph{So,} / \emph{It
follows that} / \emph{Conclusion:}) and which term the rule mentions first.

\begin{table}[H]
\caption{The $14$ template families, grouped by how the rule is formulated.
$\dagger$ marks the three families held out from all optimization. Only the
rule is shown; the fact and conclusion follow the argument form of
Table~\ref{tab:forms}. The two event families use tensed verbs rather than
adjectival predication.}
\label{tab:families}
\centering\small
\begin{tabularx}{\textwidth}{@{}llX@{}}
\toprule
Family & Formulation & Rule, as stated (affirmative / negated) \\
\midrule
\texttt{if\_then}          & material conditional & If the cube is little, then it is yellow. / \dots then it is not yellow. \\
\texttt{fact\_first}       & material conditional & \emph{(fact stated first)} If the cube is little, then it is yellow. \\
\texttt{reversed\_if}      & material conditional & The cube is yellow if it is little. / The cube is not yellow if\dots \\
\texttt{suppose\_observe}$^\dagger$ & material conditional & Suppose that if the cube is little, then it is yellow. \\
\midrule
\texttt{whenever}$^\dagger$ & generic / habitual & Whenever the cube is little, it is yellow. / \dots it is not yellow. \\
\texttt{reversed\_whenever} & generic / habitual & The cube is yellow whenever it is little. \\
\texttt{always\_generic}    & generic / habitual & Cubes that are little are always yellow. / \dots are never yellow. \\
\midrule
\texttt{all\_are}          & universal quantification & All trains are clean. / No trains are clean. \\
\texttt{all\_nouns}        & universal quantification & All trains are bolts. / No trains are bolts. \\
\midrule
\texttt{subset}$^\dagger$  & set-theoretic & The set of trains is a subset of the set of bolts. / \dots is disjoint from\dots \\
\midrule
\texttt{if\_then\_event}   & temporal, over events & If the shelf whistles, then the coin glows. \\
\texttt{every\_time\_event}& temporal, over events & Every time the shelf whistles, the coin glows. \\
\midrule
\texttt{guarantee}         & sufficiency predicate & Being little guarantees that the cube is yellow. \\
\midrule
\texttt{semi\_formal}      & symbolic & Rule: if P then W. / Rule: if P then not W. \\
\bottomrule
\end{tabularx}
\end{table}

\subsection{Control tasks}
\label{app:controls}

\paragraph{Word recall (verbal working memory).} A list of concrete nouns is
presented, followed by a two-way choice between one word that appeared and one
that did not. No inference is required, only maintenance and matching. Easy
items use an 8-word list, hard items a 48-word list.

{\small\begin{verbatim}
Here is a list of words:
dolphin, tiger, urchin, mitten, hornet, kettle, barrel, summit

Which of these two words appeared in the list above? Answer with
just 1 or 2 and nothing else.
1: barrel
2: velvet
\end{verbatim}
}

\noindent Gold: \texttt{1}. Both models are at ceiling ($1.000$).

\paragraph{Language localizer (sentence vs.\ pseudoword string).} Adapted from
the sentences-versus-pseudowords contrast used to localize the fMRI language
network: is this string an English sentence or a nonsensical set of words?
Targets linguistic form rather than memory or inference.

{\small\begin{verbatim}
Please read the following text. Is it an English sentence, or a
nonsensical set of words? Answer 1 if it is an English sentence, or
2 if it is a nonsensical set of words. Answer with just 1 or 2 and
nothing else.

"Kake wews baps ose recossed ree lenchen wesen ef ree susser
prodenotions."
\end{verbatim}
}

\noindent Gold: \texttt{2}. Both models are at ceiling ($1.000$).

\paragraph{PrOntoQA (multi-step deduction over invented categories).} A set of
universal statements over nonce categories, then a query about a named
individual. Every queried category is mentioned in the premises, so mere
presence in the context carries no information; positives vary in hop depth
from one to three, and negatives are either converse errors---the same fallacy
as AC and DA above---or unrelated stated branches. Premise order is shuffled so
position does not leak the answer.

{\small\begin{verbatim}
Answer 1 if the statement is true, or 2 if it is false. Answer with
just 1 or 2 and nothing else.
All cumps are plints. All grixs are cumps. All defots are corples.
All snicks are cumps. All blints are snicks. Hannah is a blint.
Is Hannah a snick?
\end{verbatim}
}

\noindent Gold: \texttt{1} (one hop). A converse-error negative has the same
shape but queries a category the individual is not stated to belong to, e.g.\
\emph{``\dots\ All grixs are tromps. All trants are tromps. Nate is a lemp.
Is Nate a grix?''} $\rightarrow$ \texttt{2}. This control is a
\emph{predecessor} of the present version: an earlier build drew every False
query from a category \emph{absent} from the premises, so the heuristic
``answer True iff the queried word appears'' scored $800/800$ and the
deduction chain was decorative. Qwen scores $0.998$ on that build and $0.736$
here; we report only the latter. Note the resulting ceiling difference between
models: the usable range above chance is $0.236$ for Qwen but $0.489$ for
Gemma ($0.989$ intact), so cross-model comparisons on this task must be
normalized against the intact baseline.

\paragraph{MMLU (factual knowledge).} Items are drawn from $59$ sources: $57$
MMLU subjects plus ARC-Challenge and LogiQA. Three of these are logic sources
---MMLU formal logic, MMLU logical fallacies, and LogiQA---and are held as
evaluation-only, quarantined from every optimization objective, which makes
them the sharpest generalization probes in the battery. Items are reduced to a
binary choice between the gold answer and one distractor, with side randomized.

{\small\begin{verbatim}
Answer the following question. Reply with EXACTLY ONE CHARACTER --
1 or 2 -- and nothing else. Do not write "Answer", punctuation, or
any explanation.
Which is an example of a learned behavior?
1. A salmon migrates up a river.
2. A tiger hunts a deer.
Answer:
\end{verbatim}
}

\noindent Gold: \texttt{2}. Accuracy is the unweighted mean over the $59$
sources.

\paragraph{Stroop (cognitive control).} Colour--word interference rendered in
text: the prompt states a word and the ink colour it is printed in, and the
model must report the ink colour rather than the word. Congruent items agree,
incongruent items conflict.

{\small\begin{verbatim}
You will be shown a word and the ink color it is printed in.
Respond with the ink color, not the word itself.

The word 'RED' is printed in red ink.

1. green
2. red

Reply with EXACTLY ONE CHARACTER -- 1 or 2 -- and nothing else.
\end{verbatim}
}

\noindent Gold: \texttt{2} (congruent). An incongruent item states a colour
word printed in a different ink, so the word and the correct answer diverge.

\paragraph{MSIT (multi-source interference).} Three digits are shown, exactly
one differs from the other two, and the model reports the value of the odd one.
This task uses \texttt{A}/\texttt{B} labels by necessity: its native answers are
the digits $1$--$3$, so \texttt{1}/\texttt{2} labels would collide with the
answer space and a reply of ``1'' would be ambiguous between the label and the
digit.

{\small\begin{verbatim}
Three digits are shown below. Exactly one digit differs from the
other two. Respond with the value of the differing digit.

Digits: 1 0 0

A. 2
B. 1

Reply with EXACTLY ONE CHARACTER -- A or B -- and nothing else.
\end{verbatim}
}

\noindent Gold: \texttt{B}. Both models are at ceiling ($1.000$).

\paragraph{Raven (abstract reasoning).} Raven-style $3\times3$ matrix
completion over digit patterns with the bottom-right cell blank, reduced from
four options to a binary choice.

{\small\begin{verbatim}
Please look at the following 3x3 matrix of digit patterns. Each cell
is shown in brackets; the bottom-right cell is blank.

[1] [1] [1]
[0] [0] [0]
[4] [4] [_]

Which option completes the blank cell?

1. [4]
2. [1]

Reply with EXACTLY ONE CHARACTER -- 1 or 2 -- and nothing else.
\end{verbatim}
}

\noindent Gold: \texttt{1}. For Qwen this column is not interpretable: intact
accuracy is $0.540$ against a chance line of $0.500$, so the model cannot
really do the task and a lesioned score should not be read as evidence in
either direction. Gemma is at $0.692$, which does leave headroom, so the
caveat is model-specific rather than a property of the task.

\paragraph{Multiplication (arithmetic).} One- and two-digit products answered
with the number itself rather than a label. This task is deliberately left in
its native format, and is the one departure from first-character scoring: most
two-digit products are not single tokens, so the reply is read as the leading
number of an eight-token generation and compared exactly. We report the
one-digit set, on which both models are at ceiling; on the two-digit set Qwen
is at $0.500$, which is as uninterpretable as Raven and is not used.

{\small\begin{verbatim}
Compute the following product. Answer with the number alone as the
first word of your answer.

4 * 9 =
\end{verbatim}
}

\noindent Gold: \texttt{36}.

\paragraph{Language-model continuation.} The model continues a WikiText-2
passage of more than $90$ words. Scored as the mean per-token KL divergence
between the clean and lesioned next-token distributions, so $0$ means identical
and larger is worse; there is no accuracy and no chance level. This control
catches lesions that preserve a forced-choice readout while destroying the
model as a language model.

{\small\begin{verbatim}
Continue the following text. Reply with the continuation only,
nothing else.

Robert Boulter is an English film , television and theatre actor .
He had a guest @-@ starring role on the television series The Bill
in 2000 . This was followed by a starring role in the play Herons
written
\end{verbatim}
}

\paragraph{Accuracy and response format.}
Accuracy measures the fraction of responses matching the
gold answer. Responses outside the task's allowed answer
set, such as \texttt{1}/\texttt{2} or \texttt{A}/\texttt{B},
are scored as incorrect.

\subsection{Output-channel control}
\label{app:channel}

Because all rewritten tasks share one response channel, a subspace could
preserve the channel rather than the task. We therefore evaluate the same MMLU
items in both \texttt{1}/\texttt{2} and \texttt{A}/\texttt{B} label spaces. A
gap between them measures carriage of the output channel rather than of task
content.

{\small\begin{verbatim}
Answer the following question. Reply with EXACTLY ONE CHARACTER --
A or B -- and nothing else. Do not write "Answer", punctuation, or
any explanation.
Which is an example of a learned behavior?
A. A salmon migrates up a river.
B. A tiger hunts a deer.
Answer:
\end{verbatim}
}

\noindent Gold: \texttt{B}. The two label spaces are interchangeable before
intervention: Qwen scores $0.799$ in \texttt{A}/\texttt{B} against $0.805$ in
\texttt{1}/\texttt{2}, and Gemma $0.857$ against $0.851$---within $0.006$ for
both models, with the sign of the difference reversing between them. Two
further controls point the same way: the instructed four-way MMLU format
($0.741$ intact for Qwen), and multiplication, which is answered with the
number itself and so does not use the shared channel at all.

\section{Implementation details for MVS}
\label{app:optim}

\paragraph{Operational definition.}
Let $a_\tau(U)$ denote held-out task accuracy when retaining the
subspace spanned by $U$, $a_\tau^0$ the unmodified model's accuracy,
and $c_\tau$ the chance baseline, with $a_\tau^0>c_\tau$. Define
recovery as
\begin{equation}
    \operatorname{Rec}_\tau(U)
    = \frac{a_\tau(U)-c_\tau}{a_\tau^0-c_\tau}.
\end{equation}
For any chosen recovery threshold $\rho\in[0,1]$, the smallest
sufficient rank found is
\begin{equation}
    \widehat{k}_{\ell,\tau}(\rho)
    =
    \min_{U\in\mathcal{C}_{\ell,\tau}}
    \left\{
        \operatorname{rank}(U):
        \operatorname{Rec}_\tau(U)\geq\rho
    \right\},
\end{equation}
where $\mathcal{C}_{\ell,\tau}$ contains the fitted subspaces across
the tested ranks and optimization runs. An MVS at recovery level
$\rho$ is any qualifying subspace attaining this rank. Recovery of
$1$ corresponds to the unmodified model's accuracy, and $0$ to
chance. If no fitted subspace reaches the threshold, we report
that none was found within the tested ranks. Because optimization
may miss smaller solutions, the rank found is an upper bound on
the true minimum under this criterion.

\paragraph{Initialization.} $V_0$ is drawn inside the top-$m$ principal
subspace of the layer's own $\mu$-centred activation covariance
($m=2048$): $V_0=Q_m\xi$ with $\xi_{ij}\sim\mathcal{N}(0,1/m)$ and $Q_m$ the
leading eigenvectors. A plain Gaussian init leaves the optimizer at $\ln 2$ at
most layers; the on-manifold init escapes it.

\paragraph{Reference activation.} $\mu_\ell$ is the mean of $h_\ell$ over all
token positions of a neutral corpus that appears in no evaluated task, so it
encodes no task in particular. Both operators preserve $\mu_\ell$ exactly.

\paragraph{Ablation twins.} Each ablation condition reuses the retention
solution with the projector complemented, i.e. $I-UU^T$ with the
$U$ obtained by minimizing \eqref{eq:objective}. No separate optimization is
performed for ablation.

\section{An optimizer pathology from large-magnitude coordinates}
\label{app:massive}

In Qwen3-8B, a single residual-stream coordinate (unit $2276$)
switches on at layer $6$ and carries $99.98\%$ of the
$\mu$-centred variance from there onward, with
$\lambda_1/\lambda_2$ rising from $21.6$ at layer $2$ to
$8.5\times10^{4}$ at layer $6$. A second coordinate (unit $233$)
is $830\times$ the median unit magnitude. Both are concentrated
at attention-sink positions: $56\%$ of their squared mass sits
on the first token, which accounts for $1.2\%$ of positions.

These large-magnitude coordinates are associated with an
optimization pathology: without additional constraints,
gradient descent stalls near $\ln 2$ at layers $6$--$14$
across the tested ranks. Successful late-layer subspaces
have only chance-level overlap with these coordinates while
still reproducing the clean distribution, indicating that
their large variance does not make them indispensable
at every layer.

To address the optimization failure, we fix $m$ coordinate
directions within the retained subspace and optimize the
remaining $k-m$ directions in their orthogonal complement.
Crucially, $k$ is always the \emph{total rank}, including all
pinned directions: pinning two coordinates in a rank-$k$
condition leaves $k-2$ directions to learn, rather than
increasing the rank to $k+2$. The same accounting applies
when one or no coordinates are pinned, and $m\leq k$.

At layer $6$, pinning two coordinates reduces the attained
KL from approximately $0.67$ to $3.4\times10^{-3}$.
Pinning one coordinate yields $0.26$; pinning four or eight
provides no further gain. Thus, failure of the unconstrained
optimizer does not establish that the tested rank is
insufficient. Gemma-4-12B has no comparable coordinate
($\lambda_1/\lambda_2\leq3.6$ at every layer) and requires
no such remedy.

For affected conditions, we report the solutions obtained
with pinned coordinates alongside the unconstrained results.
All reported ranks include the pinned coordinates.
Where the unconstrained optimizer succeeds, the two
procedures give consistent results.

\end{document}